\documentclass[letterpaper, 10 pt, conference]{ieeeconf}  

\IEEEoverridecommandlockouts                              

\usepackage{graphics} 
\usepackage{graphicx}
\usepackage{amsmath} 
\usepackage{amssymb}  
\usepackage{gensymb}

\usepackage{algorithm}
\usepackage{algorithmic}
\makeatletter
\newcommand\fs@norules{\def\@fs@cfont{\bfseries}\let\@fs@capt\floatc@ruled
  \def\@fs@pre{}%
  \def\@fs@post{}%
  \def\@fs@mid{\kern3pt}%
  \let\@fs@iftopcapt\iftrue}
\makeatother
\floatstyle{norules}
\restylefloat{algorithm}

\title{\LARGE \bf
Precise Synthetic Image and LiDAR (PreSIL) Dataset for Autonomous Vehicle Perception
}

\author{Braden Hurl$^{1}$, Krzysztof Czarnecki$^{2}$, and Steven Waslander$^{3}$
\thanks{$^{1}$Braden Hurl is with the David R. Cheriton School of Computer Science,
        University of Waterloo, 200 University Avenue West, Waterloo, Ontario, Canada
        {\tt\small bdhurl@uwaterloo.ca}}%
\thanks{$^{2}$Krzysztof Czarnecki is with the Department of Electrical and Computer Engineering,
        University of Waterloo, 200 University Avenue West, Waterloo, Ontario, Canada
        {\tt\small k2czarne@uwaterloo.ca}}%
\thanks{$^{3}$Steven Waslander is with the Institute for Aerospace Studies,
        University of Toronto, 27 King's College Cir, Toronto, Ontario, Canada
        {\tt\small stevenw@utias.utoronto.ca}}%
}

\begin{document}

\maketitle
\thispagestyle{empty}
\pagestyle{empty}

\begin{abstract}

We introduce the Precise Synthetic Image and LiDAR (PreSIL) dataset for autonomous vehicle perception. Grand Theft Auto V (GTA V), a commercial video game, has a large detailed world with realistic graphics, which provides a diverse data collection environment. Existing works creating synthetic LiDAR data for autonomous driving with GTA V have not released their datasets, rely on an in-game raycasting function which represents people as cylinders, and can fail to capture vehicles past 30 metres.  Our work creates a precise LiDAR simulator within GTA V which collides with detailed models for all entities no matter the type or position. The PreSIL dataset consists of over 50,000 frames and includes high-definition images with full resolution depth information, semantic segmentation (images), point-wise segmentation (point clouds), and detailed annotations for all vehicles and people. Collecting additional data with our framework is entirely automatic and requires no human annotation of any kind. We demonstrate the effectiveness of our dataset by showing an improvement of up to 5\% average precision on the KITTI 3D Object Detection benchmark challenge when state-of-the-art 3D object detection networks are pre-trained with our data. The data and code are available at https://tinyurl.com/y3tb9sxy

\end{abstract}

\section{INTRODUCTION}

Dynamic object detection is a fundamental capability required for autonomous vehicle navigation and robot operation in dynamic environments. Significant progress has been made on 2D object detection in the last decade, with seminal works such as AlexNet \cite{alexnet} and ResNet \cite{resnet} leading the way. On the KITTI Object Detection Benchmark for autonomous driving \cite{kitti}, 2D object detection methods that identify bounding boxes of objects in the image plane are now consistently over 90\% average precision (AP). For 3D autonomous vehicle perception challenges, which must identify the position and extend an object bounding box in three dimensions, LiDAR (Light Detection and Ranging) point clouds are commonly used as a source of depth information to augment image data. Training deep learning methods on large, sparse, unordered point clouds is challenging and there remains a large gap between 2D and 3D object detection scores. 

With more publicly available training data this gap could be narrowed. The data that is publicly available has large class-imbalances; the KITTI 3D Object detection training dataset consists of 28,742 cars and only 4,487 pedestrians and 1,627 cyclists. This significant class imbalance is likely playing a role in the disparity between car and pedestrian 3D detection performance, as the difference between top results on the two classes is over 30 percent AP.

For modern deep learning methods, more data typically leads to increased performance. The lack of large publicly available datasets is likely a result of the resource intensive process of annotating 3D point clouds from LiDAR scanners. Bounding and orienting objects in 3D space is a difficult and time-consuming task; the SUN RGB-D dataset \cite{sunrgbd} reported an average of 114 seconds of annotation time per 3D object instance. Lee et al. \cite{detectionlabelling} use pre-trained models to help speed up labelling of 3D object detection datasets for autonomous driving. They reduce the annotation process time by almost 95 percent with only a slight reduction in quality, but the dearth of large, publicly available datasets persists. Quality can also be a problem for human-annotated datasets; as the annotation difficulty increases, human error can also increase. Some tasks, such as point-wise segmentation, are almost beyond the capacity of humans to annotate. Furthermore, detailed class annotations are often disregarded; trucks, busses, motorbikes, and trailers are a few of the classes which are missing from the KITTI dataset.

\begin{figure}[t]
  \centering
  \vspace{1.5mm}
  \framebox{\includegraphics[width=.23\textwidth]{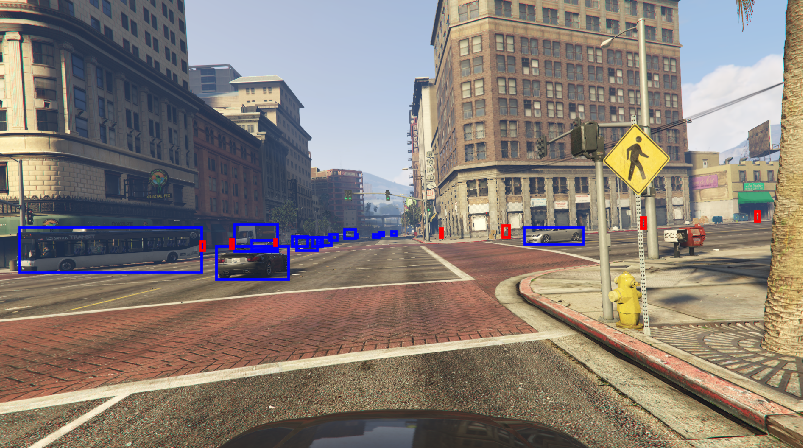}
  \includegraphics[width=.23\textwidth]{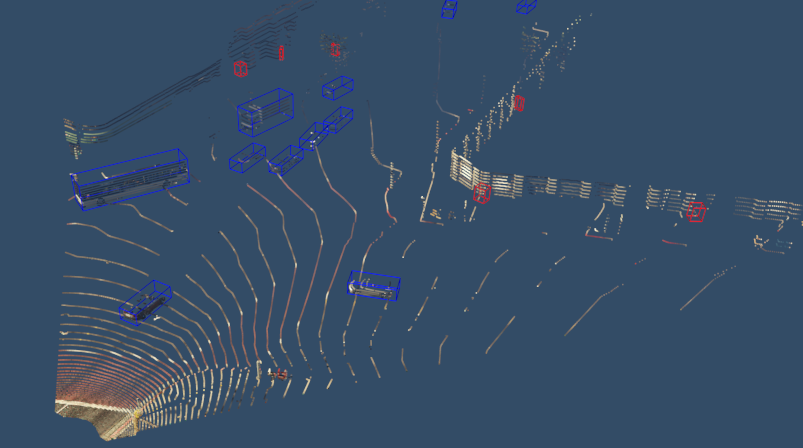}}
  
  \caption{Annotated image and point cloud pair from the PreSIL dataset}
  \label{exampleInstance}
  \vspace{-1.5mm}
\end{figure}

Synthetic data (data collected from a source other than the real world) has been used as an inexpensive means to augment real-world datasets. For example, the URSA dataset \cite{ursa}, Driving in the Matrix \cite{driveInMatrix}, and Yue et al. \cite{gtaLidar} have used the video game Grand Theft Auto V (GTA V) to produce synthetic datasets for various autonomous driving tasks. GTA V is selected for synthetic autonomous driving datasets due to the availability of numerous detailed object models, its large world, realistic graphics, and large modding (changing game source-code) community which facilitates the development of data generation tools. Once a synthetic data capturing process is established, annotated data can be generated in a fraction of the time and cost required for manual annotation. Furthermore, synthetically generated data opens up possibilities for annotations which are nearly impossible for human annotators to achieve.

We have created a process for generating perception data for autonomous vehicles with precise LiDAR and image data in GTA V. Our dataset consists of both 2D and 3D labels for object detection. We also include object segmentation for both 2D images and 3D point clouds. To summarize, our main contributions are:
\begin{itemize}
    \item We provide a novel method to generate precise LiDAR point clouds which accurately represent people (pedestrians and cyclists) using the video game GTA V.
    \item We provide a large (50,000+ frames) dataset in the KITTI format with extended information and labels. For each data frame, the dataset contains:
    \begin{itemize}
        \item A 1920 x 1080 resolution color image.
        \item A 1920 x 1080 resolution depth map.
        \item For each object in the image:
        \begin{itemize}
            \item A label in the KITTI 3D Object Detection format (2D and 3D bounding boxes, occlusion, truncation, and class information).
            \item An augmented label with entity ID, the number of 2D points in the image associated with the object, speed, pitch, roll, and a model name.
        \end{itemize}
        \item A 1920 x 1080 instance segmentation image for vehicles and people (corresponding to entity ID in label), semantic segmentation for all other pixels.
        \item An instantaneous point cloud without any motion distortion in a forward-facing 90 degree FoV. Similar to the KITTI format but without reflectance values. For each point there is also an associated entity ID if the point corresponds to a vehicle or person.
        \item Intrinsic and extrinsic calibration information for all sensors.
    \end{itemize}
    \item We demonstrate that our data can boost the performance of a state-of-the-art object detection network on the KITTI 3D Object Detection benchmark challenge \cite{kitti}.
\end{itemize}


\section{Related Work}\label{section_relatedWork}
Many works have already successfully used GTA V for generating synthetic data relating to various autonomous driving perception tasks.  In this section, we will outline related works in various domains of autonomous driving perception.

\subsection{2D Object Detection}
Driving in the Matrix \cite{driveInMatrix} focuses on 2D object detection using GTA V. The 2D bounding boxes from GTA V are produced from 3D bounding boxes transformed to the 2D perspective and are thus loose-fitting. The authors describe a method to capture depth and stencil (class-wise segmentation) buffers from the graphics pipeline. They use the stencil buffers to tighten 2D bounding boxes around classes and use the depth information to separate instances from the same stencil class. With this method they are able to obtain tight 2D bounding boxes for over 200,000 images. They show how they have obtained a larger coverage of data than Cityscapes \cite{cityscapes}. When only training on their synthetic data, they receive better results on KITTI than when only training on Cityscapes. Martinez et al. \cite{pve} also use 2D images from GTA V to detect distance to objects. They generate over 480,000 instances to train a CNN with 2D images for various tasks such as vehicle distance and lane markings. To summarize, 2D data from GTA V has provided notable performance improvements for training detection networks.

\subsection{3D point cloud Segmentation}
Yue et al. \cite{gtaLidar} and SqueezeSeg \cite{squeezeSeg} have leveraged 3D data from GTA V by creating an in-game LiDAR using ray casting. Ray casting can be done in-game through a GTA V native function. The benefit of using GTA V for point cloud segmentation is that each point has an accurate instance level segmentation annotation for any object class (i.e. vehicles, pedestrians, and cyclists). This is a difficult and time-consuming task for a human annotator to achieve as each 3D point needs to be labelled (there can be over a million points per LiDAR scan). In these works, for the KITTI data, individual points are annotated with instance level object information using 3D bounding boxes since the true annotations are unavailable. Augmenting the KITTI data (without reflectance) with GTA V point clouds increased accuracy by about 9 percent IoU.  These experiments were only conducted on cars.

Ray casting may seem like a great solution for creating point clouds in GTA V but there are major downsides with the underlying function mechanisms. SqueezeSeg \cite{squeezeSeg} notes that ``GTA-V uses very simple physical models for pedestrians, often reducing people to cylinders." Furthermore, some vehicles are simply not hit by ray casting, which could be a major problem for image and LiDAR fusion methods. Our work addresses both of these limitations.

Fang et al. \cite{baiduPCSegmentation} use deep learning to remove object points from real-world point clouds then insert simulated points from 3D object models. Using this process they are able to augment real-world data with new training samples containing different objects in the same environment without having to collect or annotate more data. They increase accuracy by adding these augmentations to their training dataset. This method does not create images that match, which is an important data element for sensor fusion methods.

\subsection{3D Scene Understanding}
Playing for Data \cite{playForData} uses GTA V to generate a semantic segmentation dataset comprised of 25,000 video frames. The experiments show that synthetic data significantly increases accuracy when supplementing smaller real-world datasets. Furthermore, it significantly reduces the amount of time and resources for obtaining larger amounts of data. Playing for Benchmarks \cite{playForBench} generates 250,000 video frames of semantic segmentation information and includes annotations for optical flow, instance-level segmentation, 2D object detection, and tracking. However, these methods use a technique called detouring, which avoids interacting with GTA V code and only allows coarse semantic segmentation labelling.

The URSA dataset \cite{ursa} increases the quality of semantic segmentation labels generated from GTA V by labelling each texture which is encountered in the game. This allows unlimited data generation without any extra annotation time. They also increase the size of the released dataset to contain over 1,000,000 images. They show further increases compared to Playing for Benchmarks and demonstrate how more synthetic data can increase network performance.

\subsection{Simulators}
Another way to generate synthetic data is through the use of simulators. SYNTHIA \cite{synthia} and CARLA \cite{carla} use commercial game-engines to create virtual worlds for autonomous vehicle simulations. Simulators are often easier to work with than video games such as GTA V since they are purpose built for autonomous driving development, their source code is often open source, and there is documentation readily available. Unfortunately, realistic worlds take large amounts of resources to develop and currently available simulators fail to reach the same level of realism that is available in commercial video games. 
CARLA has recently added LiDAR simulator capabilities with the use of ray-casting methods. However, the LiDAR simulated in CARLA is done through ray-casting onto basic collision meshes. CARLA plans to eventually release a LiDAR simulator similar to the one we have produced in GTA V\footnote{https://waffle.io/carla-simulator/carla/cards/5b9b8f959dc4a8001bdc5a87}, but it is not yet available.


\section{Synthetic Data}
To increase the usability of PreSIL for researchers, the data is formatted to be compatible with the KITTI dataset \cite{kitti}, a widely used dataset for autonomous vehicle perception. New labelling conventions are created whenever it is necessary. When there is an opportunity to create new annotations and sensory data not provided in KITTI, these are also included. Unfortunately, due to the limitations of what is possible (either through functionality or knowledge of the functionality) in GTA V, some data elements need to be approximated. This section details the capturing process, challenges encountered, limitations of the tools provided, and the format of new data.

\subsection{Capturing Process}
Our data capturing process is based on DeepGTAV's approach\footnote{https://github.com/aitorzip/DeepGTAV} for creating a scenario, driving autonomously, and retrieving object information. We also use the native code section of GTAVisionExport\footnote{https://github.com/umautobots/GTAVisionExport} for collecting depth and stencil information. Both of these repositories use Script Hook V\footnote{http://www.dev-c.com/gtav/scripthookv/} to interact with GTA V.

The camera is placed in the same position as the LiDAR scanner. This is not consistent with a realistic sensor configuration but is necessary when obtaining depth information for LiDAR point clouds as the depth buffer is generated from the camera perspective. The vehicle used is the ``ingot" model; a station wagon which closely resembles the Volkswagon Passat used in the KITTI dataset. The position of the in-game camera and LiDAR is set to match the positioning of the Velodyne LiDAR scanner from the KITTI setup \cite{kittiData} as closely as possible.

Position, dimensions, heading, type, and vehicle model are obtained directly from native in-game functions which have been documented by the GTA V modding community. 2D bounding boxes can be obtained by transforming the corners of the 3D object to the 2D image plane. This provides loose 2D bounding boxes, which can be tightened using instance-wise semantic segmentation information. Pixels are segmented in a two-step process. First by projecting the associated depth pixel into 3D space and calculating if it is situated in a 3D bounding box of the same class as the stencil pixel. If it is not, or it falls into multiple 3D bounding boxes we use a process similar to Driving in the Matrix \cite{driveInMatrix}, which uses the stencil buffer to segment object classes then depth disparity to separate instances.

\subsection{Precise LiDAR point clouds}
Previous LiDAR point cloud generation methods in GTA V have used the native ray casting function. As mentioned in section \ref{section_relatedWork}, this function uses collision meshes which are simplified models (e.g. cylinders for pedestrians). SqueezeSeg \cite{squeezeSeg} has noted this and avoided using pedestrians for this reason. Furthermore, some objects are not hit with the ray casting function beyond approximately 30 metres. User-made documentation notes that ray casting ``will not register for far away entities. The range seems to be about 30 metres."\footnote{http://dev-c.com/nativedb/}

\begin{figure}[t]
  \centering
  \vspace{1.5mm}
  \framebox{{\includegraphics[width=0.23\textwidth]{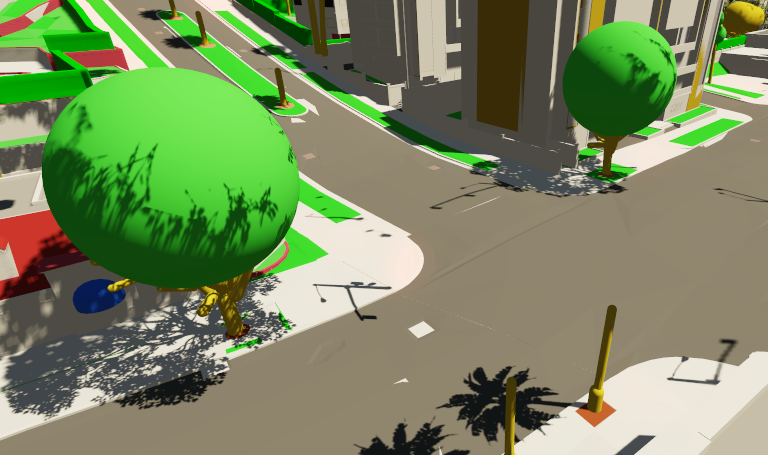}
             \includegraphics[width=0.23\textwidth]{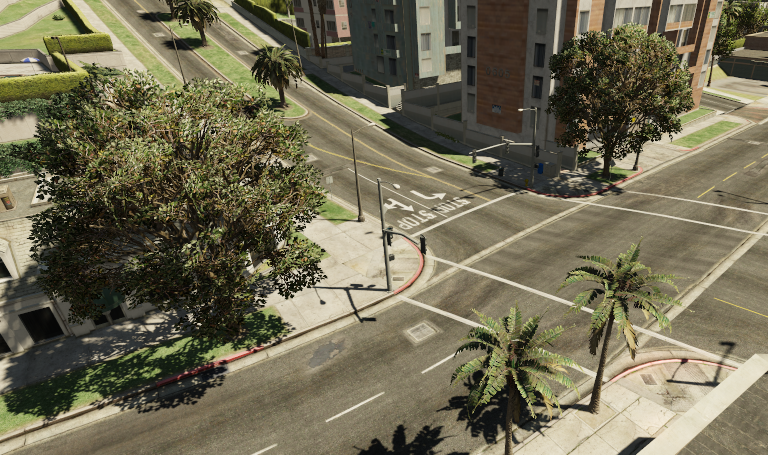}}}
  
  \caption{Collision meshes simplify physics calculations. A screenshot from GTA V shows the difference between the collision meshes (left) and the final rendering (right).}
  \label{collisionMeshes}
  \vspace{-1.5mm}
\end{figure}

It is possible to observe the details of collision meshes in Codewalker, a GTA V mod. Fig. \ref{collisionMeshes} shows that the collision meshes (used for simplifying physics calculations) of some models are not as accurate as the rendered image portrays.\footnote{https://www.gta5-mods.com/tools/codewalker-gtav-interactive-3d-map} Unlike ray casting, the depth buffer is generated using accurate models which can be used to create a point cloud generation process with precise depth information.

A point cloud $P$ is defined as a set of $n$ points, $p_i \in P = (x_i,y_i,z_i) \in \mathbb{R}^3, i = \{1, \ldots, n\}$. The coordinate system from KITTI is used where $x$ is right, $y$ is down, and $z$ is forward (with $\hat{\mathbf{i}}$, $\hat{\mathbf{j}}$, $\hat{\mathbf{k}}$ the respective unit vectors). An image, $I$ is an array of pixels at locations $(u,v)$ with associated RGB values $(R_{uv},G_{uv},B_{uv})$. A depth buffer, $D$ is an array of pixels at locations $(u,v)$ with associated depth buffer values $D_{uv}$, where the value at $(u,v)$ corresponds to the same pixel in the image $I$. The depth buffer $D$ is stored in a non-linear format. A transformation function, denoted $T_D$, is required to go from the depth buffer format to a depth value $d_{uv} = T_D(D_{uv})$. The function $T_D(D_{uv})$ is detailed in section \ref{subsec_depthcalc}.

An instance segmentation image, $S$ is an array of pixels at locations $(u,v)$, each with an associated entity ID. Each point $p_i$ is also given an associated entity ID $e_i$ if the point corresponds to an object of interest (e.g. pedestrians, vehicles). A projection transformation function, denoted $T_P$, is required to go from a vector in $\mathbb{R}^3$ to a value $(u,v)$ corresponding to a point on $I$, $D$, and $S$. $T_P$ uses camera and in-game parameters and is described in detail by Racinsky \cite{racinsky}. A transformation using $T_P$ gives decimal $u$ and $v$ values. The resulting point $(u,v)$ is located between four pixels which form a square. If ceil and floor represent rounding up and down to the nearest integers respectively, then the square of four pixels surrounding the point $(u,v)$ are given by (ceil(u),ceil(v)), (floor(u),ceil(v)), (ceil(u),floor(v)), (floor(u),floor(v)). Let $(u_1,v_1)$ be the closest pixel which is a corner of the square and $(u_4,v_4)$ be the furthest of these pixels and $GetNear(u,v)$ be a function to obtain these pixels.

To generate a point cloud, unit vectors $\vec{r}_{\theta\phi}$ are created at various horizontal ($\theta$) and vertical ($\phi$) angular resolutions to simulate a Velodyne HDL-64E. Each unit vector is projected to the image plane and a depth value is obtained through an interpolation process. This method relies on the LiDAR and camera being located in the same position. Horizontal angular resolution $\theta_r$ of 0.09 degrees is used when rotating around the $y$ (down) axis of the vehicle. The minimum and maximum horizontal angles are defined as $\theta_{min} = -45\degree$ and $\theta_{max}=45\degree$. Vertical angular resolution $\phi_r$ of 0.42 degrees is used corresponding to a rotation around the $x$ (right) axis. Minimum and maximum vertical angles are defined as $\phi_{min}=-24.9\degree$ and $\phi_{max}=2\degree$, which corresponds to 64 vertical beams. Let $d_{max}$ be the maximum range (120 metres for the Velodyne HDL-64E). Let $R(\theta) \in SO(3)$ be a rotation of $\theta$ degrees about the $y$-axis. Let $R(\phi) \in SO(3)$ be a rotation of $\phi$ degrees about the $x$-axis.

Algorithm \ref{pcAlgo} defines the PreSIL process of generating a point cloud. Without interpolation the point cloud consists of straight lines of points. By performing step \ref{threshStep} it eliminates points with large disparity being interpolated which can cause artifacts (points in between objects). The ratio of 1.08 was determined through visual inspection of point clouds.

\begin{figure}[htbp]
  \vspace{1.5mm}
  \centering
  \framebox{{\includegraphics[width=0.23\textwidth]{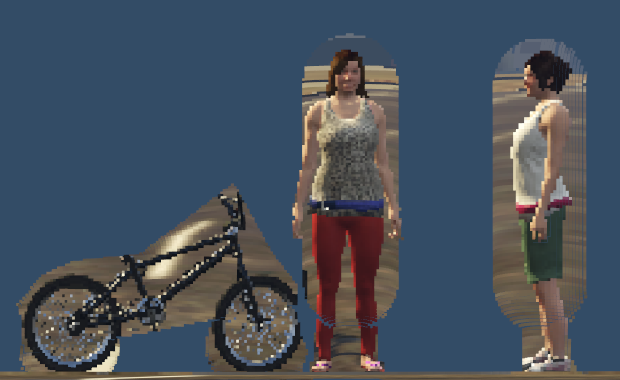}
             \includegraphics[width=0.23\textwidth]{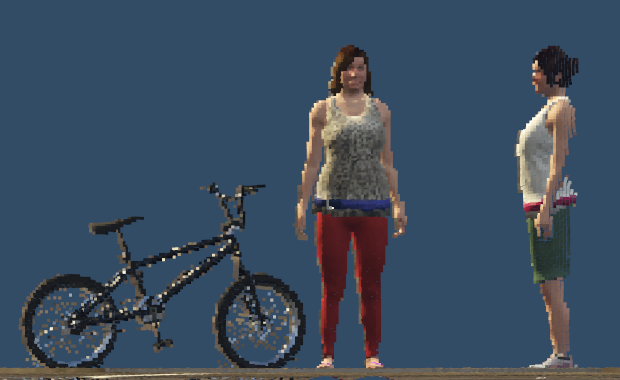}}}
  
  \caption{Image colors projected onto a point cloud display the difference between ray casting (left) and the PreSIL point cloud generation method (right) with a view of pedestrians and a bicycle. Note: For display purposes a higher LiDAR resolution was used for this graphic.}
  \label{detailedPeds}
\end{figure}

\begin{algorithm}[ht]
\caption{Algorithm for generating a point cloud}
\begin{algorithmic}[1]\label{pcAlgo}
\STATE $i\gets 0$
\FOR {$\theta | \theta_{min} < \theta < \theta_{max}, \theta \mod \theta_r = 0$}
    \FOR {$\phi | \phi_{min} \leq \phi \leq \phi_{max}, \phi \mod \phi_r = 0$}
        \STATE $\vec{r}_{\theta\phi} \gets R(\theta) R(\phi) \hat{\mathbf{k}}$
        \STATE $u,v\gets T_P(\vec{r}_{\theta\phi})$
        \STATE $(u_j,v_j) | 1 \leq j \leq 4 \gets GetNear(u,v)$
        \FOR{$j | 1 \leq j \leq 4$}
            \STATE $d_j \gets d(u_j,v_j)$
        \ENDFOR
        \IF {$\max(d_1,\ldots,d_4) < 1.08 \cdot \min(d_1,\ldots,d_4)$}\label{threshStep}
            \STATE $d_r\gets BilinearInterpolation(d_1,...,d_4)$
        \ELSE
            \STATE $d_r\gets$ $d_1$
        \ENDIF
            
        \IF {$d_r < d_{max}$}
            \STATE $p \gets \vec{r}_{\theta\phi} \cdot d_r$
            \STATE $e_i \gets S_{(u_1,v_1)}$
            \STATE $p_i \gets \{p, e_i\}$
            \STATE $P \gets P \cap p_i$
            \STATE $i\gets i + 1$
        \ENDIF
    \ENDFOR
\ENDFOR
\STATE $n \gets i$
\end{algorithmic}
\end{algorithm}

\begin{figure}[htbp]
  \vspace{1.5mm}
  \centering
  \framebox{{\includegraphics[width=0.23\textwidth]{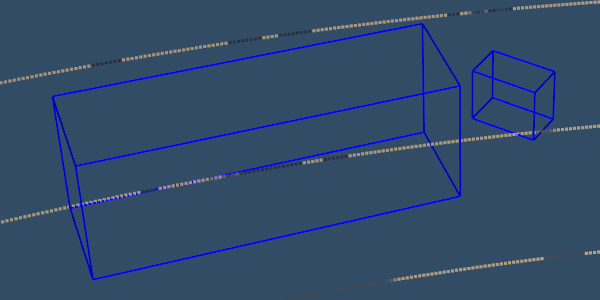}
             \includegraphics[width=0.23\textwidth]{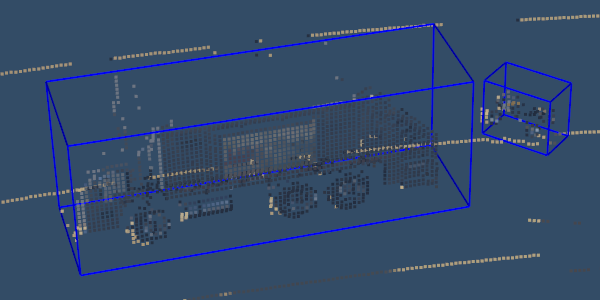}}}
  
  \caption{Comparison between point clouds generated using ray casting (left) and our method (right). Some vehicles past 30 metres are not hit with ray casting but are captured with our method.}
  \label{vehiclesDisappearing}
\end{figure}

This results in precise point clouds which accurately represent all object classes. A comparison of a point cloud generated using ray casting and our method is displayed in Fig. \ref{detailedPeds}. The image focuses on pedestrians and bicycles as these two classes are drastically improved by our method. Fig. \ref{vehiclesDisappearing} demonstrates how a ray casting point cloud may not hit all objects and, captured simultaneously, shows how our method hits the vehicles which ray casting missed.

\begin{figure}[htbp]
  \centering
  \framebox{{\includegraphics[width=0.23\textwidth]{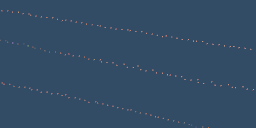}
             \includegraphics[width=0.23\textwidth]{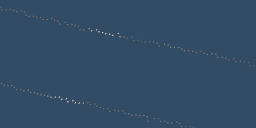}}}
  
  \caption{Similarity between depth noise in KITTI (left) and PreSIL (right).}
  \label{lidarNoise}
  \vspace{-1.5mm}
\end{figure}

Gaussian noise is added to each point consistent with the accuracy (2 centimetres) of the Velodyne HDL-64E used for the KITTI dataset \cite{kittiData}. The Gaussian Noise is added with a standard deviation of 6 millimetres, which has approximately a 99.9 percent chance of being under 2 centimetres. The similarities between the line compositions can be seen in Fig. \ref{lidarNoise}.

\subsection{Depth Calculation}\label{subsec_depthcalc}
Using information from Racinsky \cite{racinsky} as a starting point, the function $T_D$ was reverse engineered by examining object points at various distances until points associated with objects were within the 3D bounding boxes. The equation for the transformation $T_D$ to obtain a depth value $d_{uv}$ for a pixel $(u,v)$ is:

\begin{equation}
    d_{uv} = \frac{d_{nc}}{D_{uv} + \frac{nc_z \cdot d_{nc}}{2 \cdot fc_z}}
\end{equation}

Where $d_{nc}$ is the distance to the point on the near clipping plane for a pixel (u,v) on the image given by the equation:
\begin{equation}
    d_{nc} = \sqrt{nc_x^2 + nc_y^2 + nc_z^2}
\end{equation}
The variables $nc_z$ and $fc_z$ represent the near and far clipping plane distances respectively and are obtained through in-game functions. $nc_x$ and $nc_y$ are the horizontal and vertical near clipping plane distances, respectively, from the center of the clipping plane to the point $(u,v)$ on the image. These are given by the equations:
\begin{equation}
    nc_x = \mid \frac{2 \cdot u}{nc_W - 1} - 1 \mid \cdot \frac{nc_W}{2}
\end{equation}
\begin{equation}
    nc_y = \mid \frac{2 \cdot v}{nc_H - 1} - 1 \mid \cdot \frac{nc_H}{2}
\end{equation}

The near clipping plane width and height are denoted by $nc_W$ and $nc_H$ respectively. They are calculated by using the screen aspect ratio (AR), near clipping plane distance, and the vertical field of view ($fov_V$) (all of which can be obtained by in-game functions). These values remain the same unless graphics settings are changed.
\begin{equation}
    nc_H = 2 \cdot nc_z \cdot \tan(\frac{fov_V \cdot \frac{180}{\pi}}{2})
\end{equation}
\begin{equation}
    nc_W = AR \cdot nc_H
\end{equation}

\subsection{Synchronization}
Our data collection process uses both native functions and DirectX buffers to collect data. Synchronization between the data sources proved to be a major challenge. Since the source code is not available, it is impossible to know exactly how the game engine functions. The depth and stencil buffers are obtained from the DirectX buffers, whereas native functions are used to obtain information about the object and to perform ray casting. When the buffers arrive from DirectX, a callback function can be used to time the collection of all other data. However, the native functions are at a different time-step than the buffers. If buffers are used from one time-step after using the native functions, the different data sources align most of the time. However, there are some instances which still do not align. Our method of synchronization uses native functions to pause the game, flush the graphics buffers, collect synchronized data, then resume the game. This procedure is completed in the following steps:
\begin{enumerate}
    \item Pause game
    \item Two calls to render the script cameras
    \item Collect data (from native functions and DirectX)
    \item Resume the game
\end{enumerate}
This procedure has been tested on graphics processing units (GPUs) from both Nvidia and AMD and has been visually inspected to ensure point clouds align. The visual inspection was completed throughout a predetermined route during which the vehicle regularly maneuvers around corners. Synchronization and alignment problems are easily seen when the camera is rapidly rotating as it creates greater inter-frame disparity.

\subsection{Limitations}
GTA V does not contain reflectivity information. LiDAR uses light from lasers to determine the distance to a surface. It also uses this to obtain a reflectivity value which is commonly used for detecting objects. This value is omitted from our point clouds. The reflectivity value also determines the range at which a LiDAR scanner can detect a surface. If there is low reflectivity, the detectable range of a surface will be lower. This representation of reflectivity was not included in our point cloud generation method. In the future, it could potentially be incorporated by labelling reflectivity of texture values similar to how the URSA dataset \cite{ursa} labelled textures for semantic segmentation. Another potential solution could be to train a generative adversarial network (GAN) to add reflectivity information to synthetically generated point clouds.

The Velodyne HDL-64E contains a row of lasers and rotates to allow the lasers to capture data around the entire vehicle. During the KITTI data collection the scanner is rotating at 10 Hertz. This affects the point cloud in that there can be up to 100 ms difference between point capture times in the same point cloud. This is a difficult behaviour to replicate in synthetic data as there is not enough control over the passage of time in-game. For our synthetic LiDAR scanner all points are captured at the same instant in time.

Another limitation of PreSIL is that the LiDAR scanner only captures points which are in the field of view of the camera (45 degrees to either side of forward). Future work will attempt to build a 360 degree LiDAR generator using multiple cameras with different viewing perspectives.

Velodyne HDL-64E point clouds contain another source of noise where for some points a value is erroneously not returned even though the depth is within the maximum range. SqueezeSeg \cite{squeezeSeg} projects points onto a sphere to show that KITTI point clouds contain this type of noise. Our synthetic LiDAR does not attempt to approximate this source of noise.

\subsection{Extensions}
\begin{figure}[htbp]
  \centering
  \framebox{{\includegraphics[width=0.23\textwidth]{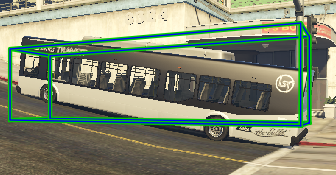}
            \includegraphics[width=0.23\textwidth]{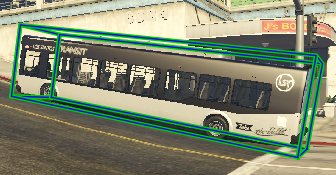}}}
  
  \caption{On steep slopes vehicles appear out of their ground truth 3D bounding boxes using the KITTI data format (left), but our augmented annotations correct this problem (right).}
  \label{vehiclePitch}
\end{figure}

While synthetic data from GTA V has some limitations, it also has a few advantages. Instead of reflectivity values for each point in the point cloud, PreSIL contains new pieces of information. First, an entity ID is provided if the point lies on an object, or the stencil class if it does not. The elevation changes also lead to further label parameters being created. As can be seen in Fig. \ref{vehiclePitch}, pitch can be necessary to complete the object information otherwise the bounding box may not encompass the entire object. Roll is also included to complement pitch.

The extended annotations are contained in a separate file to keep labels compliant with the KITTI format. For each object the augmented annotations also include:
\begin{itemize}
    \item Entity ID (to relate entity ID of points to a detection box)
    \item Count of pixels which are part of the object's 2D instance-based segmentation mask
    \item Speed
    \item The object's model string identifier
\end{itemize}

Model strings could enable interesting experiments by using vehicle models as classes instead of broader vehicle classes such as cars and trucks. Images are captured with a resolution of 1920 x 1080 pixels. Full resolution depth, stencil (class-wise segmentation), and instance-level segmentation images have also been included. These additional files are useful for semantic segmentation and depth completion tasks. Specific formatting for these files can be found in the data format document included with the dataset download.

\subsection{Dataset Statistics}

\begin{figure}[htbp]
\vspace{1.5mm}
  \centering
  \framebox{{\includegraphics[width=0.22\textwidth]{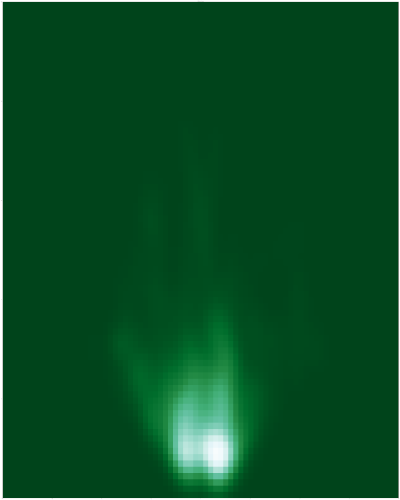}
             \includegraphics[width=0.22\textwidth]{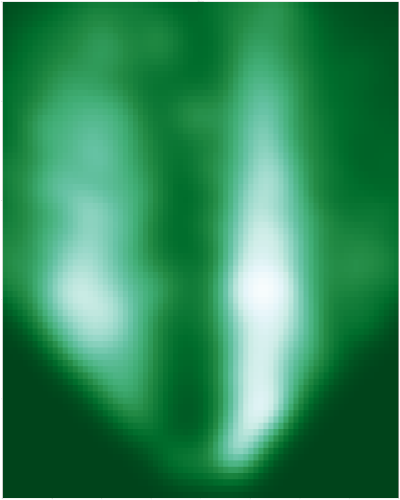}}}
  
  \caption{Birds eye view (BEV) Heatmaps of KITTI (left) and PreSIL (right) showing coverage of pedestrian locations. Heatmaps extend 40 metres to the left and right, and 100 metres forward.}
  \label{heatmaps}
\end{figure}

The count for each object class and the average count per image which contains the object class are outlined in Tab. \ref{table_overview_stats}. The PreSIL dataset has an abundance of pedestrians and vehicles but fewer cyclists compared to KITTI. The PreSIL dataset also includes important classes such as Bus, Motorbike, and Trailer, which are not included in the KITTI dataset. Fig. \ref{heatmaps} shows a bird's eye view (BEV) heatmap comparison of the KITTI and PreSIL datasets for the pedestrian class. As can be seen, the PreSIL dataset has a larger coverage area than the KITTI dataset. A large part of the discrepancy is likely due to GTA V having wider roads than the KITTI collection area. Pedestrians are encountered much further to the sides and there is a gap in the middle (the roadway) where there are fewer pedestrians in the PreSIL dataset.

\begin{table}[htbp]
\caption{Comparison of class statistics for the PreSIL and KITTI datasets where Average Per Image Containing Class (APICC) is provided to compare class densities of images which have at least one instance of the class.}
\vspace{-.5mm}
\label{table_overview_stats}
\begin{center}
\begin{tabular}{|c||c|c|c|c|}
\hline
& KITTI & PreSIL & KITTI & PreSIL\\
\hline
Type & \multicolumn{2}{c|}{Count} & \multicolumn{2}{|c|}{APICC}\\
\hline
Car & 28,742 & 334,066 & 4.30 & 7.27\\
Pedestrian & 4,487 & 147,077 & 2.52 & 6.17\\
Cyclist & 1,627 & 516 & 1.43 & 1.23\\
Truck & 1,094 & 105,316 & 1.06 & 2.90\\
Person\_sitting & 222 & 6,511 & 2.24 & 2.09\\
Motorbike & NA & 15,022 & NA & 1.37\\
Trailer & NA & 4,530 & NA & 1.20\\
Bus & NA & 7,792 & NA & 1.14\\
Railed & NA & 1,226 & NA & 3.04\\
Airplane & NA & 255 & NA & 1.00\\
Boat & NA & 606 & NA & 1.02\\
Animal & NA & 21 & NA & 1.00\\
DontCare & 11,295 & NA & 2.11 & NA\\
Misc & 973 & NA & 1.25 & NA\\
Van & 2,914 & NA & 1.36 & NA\\
Tram & 511 & NA & 1.46 & NA\\
\hline
\end{tabular}
\end{center}
\end{table}

We also added the capability to create user configurable scenes. User configurable scenes can be used to test the coverage of a model and increase the robustness by training on the data which reveals holes in the coverage. Examples of this can be seen from Yue et al. \cite{gtaLidar}. Future work may use these scenes to improve coverage of pedestrians or cyclists.


\section{Experiments and Results}
Experiments were constructed for two primary reasons:
\begin{enumerate}
    \item To prove the efficacy of our synthetic data capturing process at improving accuracy for autonomous driving perception benchmarks.
    \item To understand the impact of pre-training with synthetic data on state of the art pedestrian detection algorithms for autonomous driving.
\end{enumerate}

3D Object Detection was used for evaluation as it is a primary task for autonomous driving perception and to the best of our knowledge results for this task with synthetic and real data have never been published. We chose to use the AVOD-FPN \cite{avodfpn} architecture for testing. AVOD-FPN is a top-5 performer on the KITTI 3D object detection benchmark challenge \cite{kitti} under the pedestrian category, has publicly available code, and runs in real-time (0.1 second runtime). Pedestrians are the focus of the evaluation as there is a smaller proportion of pedestrians in the KITTI dataset and our novel LiDAR generator drastically improves the realism of pedestrians. Cyclists are not included in the experiments as there is a much smaller cyclist count in PreSIL than in the KITTI dataset. If desired, additional scenarios could be constructed where cyclists are spawned in abundance.

\subsection{Evaluation Metrics}
The trained 3D object detection network is evaluated using the standard KITTI metrics for 3D object detection. This is defined as average precision, which is simply the precision averaged over different levels of recall. The KITTI 3D object detection challenge is divided into three categories (easy, moderate, and hard) based on 2D bounding box height, occlusion, and truncation levels.

\begin{table}[htbp]
  \vspace{1.5mm}
\caption{Experimental Results: Average Precision on the Pedestrian class of the KITTI 3D Object Detection Benchmark.}
\label{table_results}
\begin{center}
\begin{tabular}{|c||c||c||c|}
\hline
Data & Moderate & Easy & Hard\\
\hline
0-4k & 51.31 \%  & 57.40 \%& 44.70 \%\\
10-4k & 51.45 \% & 57.57 \% & 47.72 \%\\
40-4k & \textbf{55.57} \% & \textbf{59.47} \% & \textbf{49.20} \%\\
40-2k & 48.04 \% & 54.14 \% & 44.62 \%\\
40-0k & 11.85 \% & 14.13 \% & 11.62 \%\\
\hline
\end{tabular}
\end{center}
  \vspace{-1.5mm}
\end{table}

\subsection{Experimental Setup}
To confirm that the PreSIL data improves the performance of state of the art object detectors, we compare the results of training and evaluating AVOD-FPN using different combinations of synthetic and real training data. The KITTI 3D Object Detection dataset is used for both training and evaluation. The KITTI 3D Object Detection training dataset is split approximately 50/50 to create which will be referred to as the KITTI training/testing splits. These splits resulted in 3,712 train instances and 3,769 test instances. Validation will not be used when training on KITTI but instead the results will be taken from the best evaluated checkpoint for each training configuration. The PreSIL data is split into an 80/20 training/validation split resulting in approximately 40,000 and 10,000 instances respectively. A subset is created of 10,000 instances from the 40,000 instance training split for use in a different training configuration.

As a baseline, AVOD-FPN is trained using only the KITTI training split. For the next set of training configurations a subset of 10,000 or the full 40,000 instances of PreSIL is used for training. First, AVOD-FPN is trained on the PreSIL train subset then evaluated with the PreSIL validation split. Next, fine-tuning on the KITTI set is completed by training from the highest performing checkpoint from the PreSIL training. Finally, all checkpoints are evaluated using the KITTI test split. The results shown are from the single best-performing checkpoint of each configuration.

As a secondary experiment AVOD-FPN is trained with subsets of the real data to show the effectiveness of synthetic data in the presence of less real data. Training configurations with half (approximately 2,000 instances) and no real-world training data are completed.

\subsection{Experimental Results}
Results are displayed in Tab. \ref{table_results}. The notation `x-y' is used to specify the number of PreSIL and KITTI data instances (in thousands) used to train AVOD-FPN. For example, 40-4k signifies 40,000 PreSIL instances and 4,000 KITTI instances were used for training.

The results are promising as there is a significant improvement (1-5\% AP) at all difficulties when pretraining with 40,000 PreSIL instances. An interesting observation is that the improvement is largest for the hard difficulty. This could be due to the disparity between the distributions of pedestrian locations in the PreSIL and KITTI datasets. Shown in Fig. \ref{heatmaps}, PreSIL has a larger average distance than KITTI and therefore a higher probability of training data being more similar to the hard difficulty level. The results in Tab. \ref{table_results} are significantly ($\approx 10 \%$) higher than on the KITTI benchmark results page since the networks are evaluated on a test split of the training data. The training data with labels provided by KITTI is likely closer in distribution within the training set than to the testing set, and therefore results in better performance.

No comparison to the impact of other datasets on network performance is possible at this time due to their current lack of availability. Future work will include the task of collecting data from a simulation environment such as CARLA for comparing performance with the PreSIL dataset.

One downside of GTA V point clouds is that there is no reflectance value available. Typical LiDAR scanners produce x, y, and z positional values as well as a reflectance measurement. Yue et al. \cite{gtaLidar} elected to simply use depth information without a reflectance value for their synthetic LiDAR generator. When only training with the KITTI dataset they show a decrease in performance when omitting the reflectance values, a drop of approximately 8 percent Intersection over Union (IoU) in their experiments. This signifies that reflectance values play a significant role in point cloud segmentation, so it is likely also important for detection. Future work could focus on estimating reflectance values from in-game properties.


\section{CONCLUSIONS}

We have created a synthetic dataset with standard annotations for autonomous driving perception tasks at a fraction of the cost of a real-world dataset. We have given a qualitative analysis on why our data collection process is superior to any other synthetic data collection methods. We have also added new annotations. The new annotations create exciting opportunities for new ways to train perception networks for various autonomous driving tasks. Future work will include adding a tracking dataset as well as 360 degree coverage of images and LiDAR. We have demonstrated how synthetic data can be used to improve state of the art 3D object detection performance and have provided a quantitative analysis on why some difficulty levels show larger improvements. Our dataset and code is publicly available\footnote{https://uwaterloo.ca/waterloo-intelligent-systems-engineering-lab/projects/precise-synthetic-image-and-lidar-presil-dataset-autonomous} and can be used by anyone to improve the performance of their perception algorithms. Future work could also include transfer learning and methods for adapting data from one distribution to another to improve training performance. These results only scratch the surface of how synthetic data can be used to achieve more robust autonomous driving perception networks.

\addtolength{\textheight}{-12cm}   


\bibliographystyle{IEEEtran}
\bibliography{IEEEabrv,mybib}

\end{document}